\documentclass[preprint,12pt]{elsarticle}

\usepackage{lineno,hyperref}
\usepackage{amsmath,amssymb,amsfonts,epsfig}
\usepackage{graphicx} %
\usepackage{array}    %
\usepackage{booktabs} %
\usepackage{makecell} %
\usepackage{setspace} %
\usepackage{graphics}
\usepackage{graphicx}
\usepackage{epstopdf}
\usepackage{amssymb}
\usepackage{subfigure}
\usepackage{url}
\usepackage{multirow}
\usepackage{algorithmic}
\usepackage{algorithm}

\journal{Elsevier}

\begin{document}

\begin{frontmatter}

\title{Training convolutional neural networks with cheap convolutions and online distillation}


\author[mymainaddress]{Jiao Xie}
\ead{jiaoxie1990@126.com}

\author[mysecondaryaddress]{Shaohui Lin}
\ead{shaohuilin007@gmail.com}

\author[mythirdaddress]{Yichen Zhang}
\ead{ethan.zhangyc@gmail.com}

\author[mymainaddress]{Linkai Luo\corref{mycorrespondingauthor}}
\cortext[mycorrespondingauthor]{Corresponding author}
\ead{luolk@xmu.edu.cn}


\address[mymainaddress]{Department of Automation, Xiamen University, Xiamen 361005, China}
\address[mysecondaryaddress]{Department of Computer Science, National University of Singapore, 117417, Singapore}
\address[mythirdaddress]{School of Informatics, Xiamen University, Xiamen 361005, China}

\begin{abstract}
The large memory and computation consumption in convolutional neural networks (CNNs) has been one of the main barriers for deploying them on resource-limited systems. 
To this end, most cheap convolutions (\emph{e.g.}, group convolution, depth-wise convolution, and shift convolution) have recently been used for memory and computation reduction but with the specific architecture designing. 
Furthermore, it results in a low discriminability of the compressed networks by directly replacing the standard convolution with these cheap ones.
In this paper, we propose to use knowledge distillation to improve the performance of the compact student networks with cheap convolutions. In our case, the teacher is a network with the standard convolution, while the student is a simple transformation of the teacher architecture without complicated redesigning. 
In particular, we propose a novel online distillation method, which online constructs the teacher network without pre-training and conducts mutual learning between the teacher and student network, to improve the performance of the student model. Extensive experiments demonstrate that the proposed approach achieves superior performance to simultaneously reduce memory and computation overhead of cutting-edge CNNs on different datasets, including CIFAR-10/100 and ImageNet ILSVRC 2012, compared to the state-of-the-art CNN compression and acceleration methods. The codes are publicly available at https://github.com/EthanZhangYC/OD-cheap-convolution.
\end{abstract}


\end{frontmatter}


\section{Introduction}
\label{sec1}
The success of convolutional neural networks in a variety of computer vision tasks \cite{krizhevsky2012imagenet,girshick2014rich,shelhamer2017fully} has been accompanied with a significant increase of memory and computation cost. It prohibits the usage of big and memory-intensive CNNs on resource-limited devices, such as mobile phones and wearable devices. To address these issues,  several techniques have been proposed for CNN compression such as low-rank decomposition \cite{lin2018holistic,lebedev2015speeding,kim2016compression,zhang2015efficient,lin2016towards}, network pruning \cite{han2015learning,han2016deep,luo2017ThiNet,he2017channel}, parameter quantization \cite{rastegari2016xnor,courbariaux2015binaryconnect,courbariaux2016binarynet} and compact network designing \cite{howard2017mobilenets,zhang2018shufflenet,chollet2017xception}. Compact network designing has received a great deal of research focus by significantly reducing the number of operations and memory cost.

The key point in designing compact networks is to replace the over-parametric convolutional filter/operation with a compact or cheap one. For example, fire module in SqueezeNet \cite{SqueezeNet}, group convolution \cite{krizhevsky2012imagenet,zhang2018shufflenet,huang2018condensenet,ioannou2017deep}, depthwise separable convolution \cite{howard2017mobilenets,sandler2018mobilenetv2,chollet2017xception,sifre2014rigid} and shift convolution \cite{wu2018shift,chen2019all}. The compact networks achieve high performance not only by their usage of cheap convolutions, but also by the elaborate designing on network architecture. However, as shown in Table \ref{tab1}, the cheap convolutions have less generalization ability to be directly applied into the given target models (\emph{e.g.}, ResNet-56 \cite{he2016deep}, WRN-40-1 \cite{zagoruyko2016wide} and DenseNet-40-12 \cite{huang2017densely})\footnote{We retrain the models by PyTorch \cite{paszke2017automatic}, which may result in a different accuracy reported in the original references.}, which leads to low accuracy when training them from scratch. Furthermore, training the compact networks needs many trial-and-error experiments with the complicated hyperparametric settings, such as the learning rate and its decay schedule. So, given a target model, can we apply these cheap convolutions without redesigning the network architecture to accelerate and compress it while retaining the commensurate accuracy?
%

Inspired by knowledge distillation \cite{hinton2015distilling,romero2015fitnets,zagoruyko2017paying,ba2014deep}, a smaller student network can learn the knowledge from the outputs of a large pre-trained teacher network to aid in the training of the student network. By doing this, the student network is more powerful to achieve better performance, compared to training it solely on the training data. 
However, the existing knowledge distillation methods require a pre-trained teacher model and rely on multi-stage training, which are unattractive in the practical applications. 

\begin{table*}[t]
\footnotesize
\begin{center}
\caption{Results of training the network with cheap convolutions (\emph{e.g.}, group convolution, depthwise convolution and shift convolution) from scratch on CIFAR-10. Group-$a$ refers to the group convolution with setting the number of groups to $a$. M/B presents million/billion in this paper. The more details of these convolutions are described in Section \ref{ssec3_2}.}
\label{tab1}
\begin{tabular}{ccccc}
\hline
Model  & Convolution & Error \% & \#Params (M) & FLOPs (M) \\
\hline
\multirow{4}*{ResNet-56 \cite{he2016deep}} & Standard & 6.20 & 0.85 & 126.81 \\ 
& Group-16 & 8.58 & 0.16 & 23.35 \\
& Depthwise & 9.55 & 0.13 & 20.14 \\
& Shift & 10.17 & 0.10 & 15.39 \\ \hline
\multirow{4}*{WRN-40-1 \cite{zagoruyko2016wide}} & Standard & 6.46 & 0.56 & 83.02 \\ 
& Group-16 & 7.94 & 0.10 & 15.98 \\
& Depthwise & 8.69 & 0.08 & 13.87 \\
& Shift & 10.63 & 0.07 & 10.67 \\ \hline
\multirow{4}*{DenseNet-40-12 \cite{huang2017densely}} & Standard & 5.19 & 1.06 & 282.92 \\ 
& Group-12 & 5.55 & 2.15 & 383.74 \\
& Depthwise & 7.30 & 0.33 & 98.91 \\
& Shift & 8.27 & 0.25 & 79.87 \\ \hline
\end{tabular}
\end{center}
\end{table*}

\begin{figure*}[t]
\centering
  \subfigure[The original target network]{
    \includegraphics[scale = 0.5]{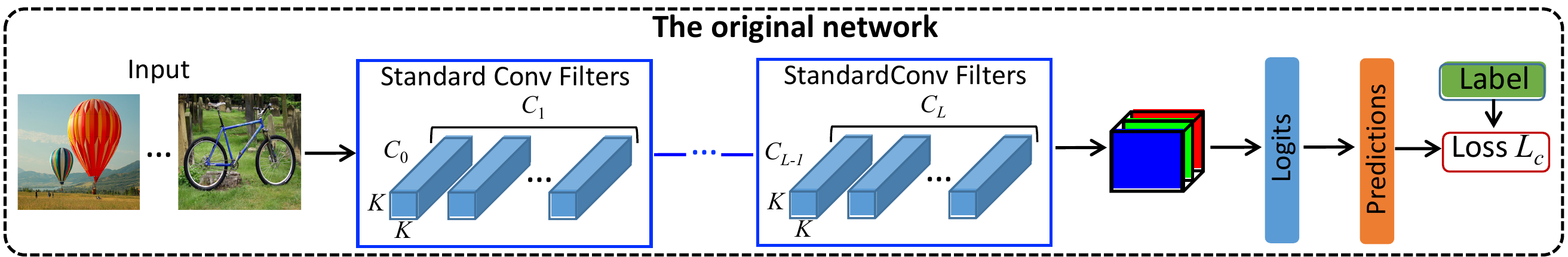}
    \label{fig1_a}
    }
  \subfigure[Online distillation with cheap convolutions]{
    \includegraphics[scale = 0.45]{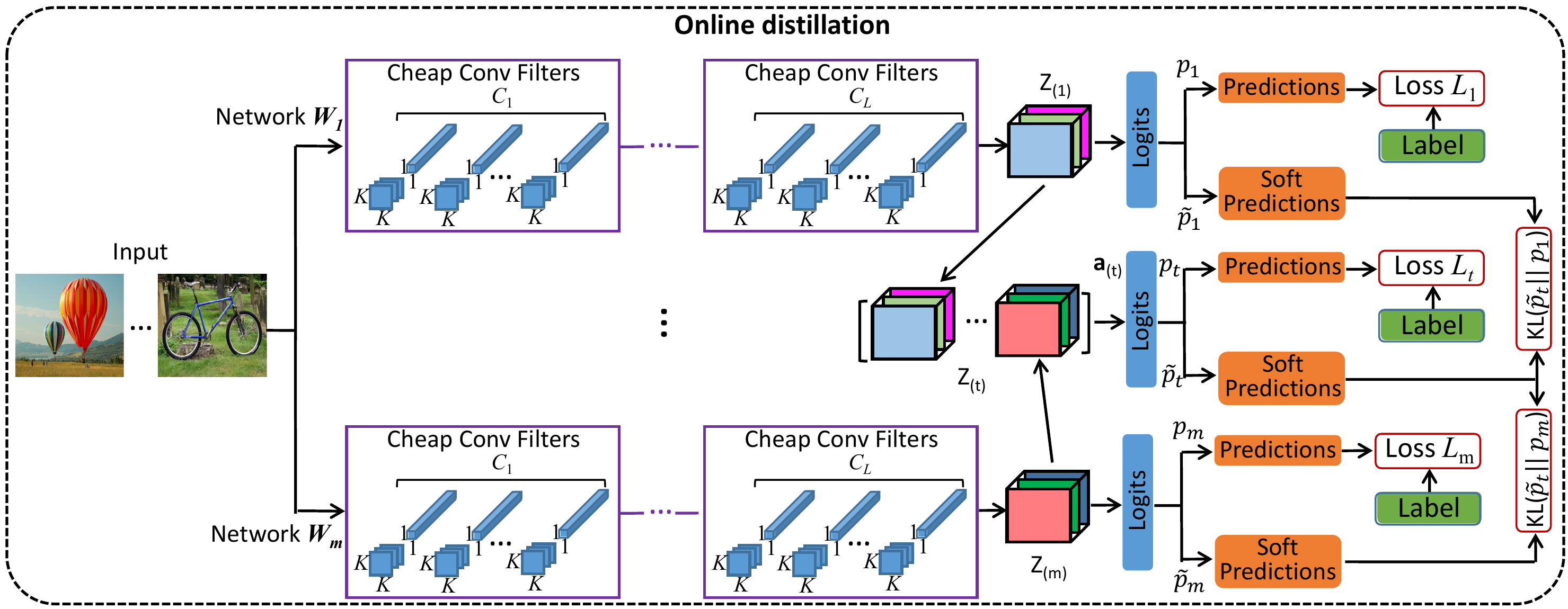}
    \label{fig1_b}
  }
\vspace{-1em}
\caption{Illustration of our online distillation method with cheap convolutions. (a) The original network with standard convolutions. (b) Online distillation is performed on the compressed network with cheap convolutions to further improve its performance. First, a student network is formed by replacing its standard convolution with the cheap ones. Then, a strong teacher network is constructed online by concatenating the output features $\mathcal{Z}_{(i)}$ from the same multiple student networks and adding the new classifier. During the training, mutual learning is conducted to improve the performance between teacher and students. For testing, the best student model in validation set is selected to be a compressed model. (Best viewed in color.) }
\label{fig1}
\end{figure*}

In this paper, instead of making student networks thinner or more shallow, we propose to use knowledge distillation to improve the performance of our compact student networks, in which the standard convolutions are only replaced with cheaper convolutions for the given original architectures. 
By doing this, we can reduce both memory and computation consumption without redesigning the student network architecture.
To get rid of the limitation of a pre-trained teacher network with multi-stage training, we further propose an \emph{online distillation} (OD) method to online generate a stronger teacher and improve the discriminative ability of the student model. Fig. \ref{fig1} shows the proposed online distillation framework.
We first replace the standard convolutions in the original network with the cheap ones to form a student network, in which the original architecture is kept. Subsequently, we online construct a strong teacher network by ensembling multiple student networks with the same architecture. In particular, the final convolutional features from each student network are extracted and concatenated to be new features of the teacher network. After that, mutual learning is proposed to improve the performance between each student model and the teacher model in a closed-loop manner. On one hand, the knowledge from the soft predictions in the teacher model is distilled back to each student to improve its discriminability. In the other hand, more high-level semantic features produced by each improved student model are generated to better improve the discriminability of the teacher model. Furthermore, different from the existing knowledge distillation methods \cite{hinton2015distilling,romero2015fitnets,zagoruyko2017paying} with the multi-stage training, Stochastic Gradient Descent (SGD) is used in OD to simultaneously train the student and teacher models in a one-shot manner.

The proposed OD is evaluated on different network architectures, including ResNets \cite{he2016deep}, DenseNets \cite{huang2017densely} and MobileNets \cite{sandler2018mobilenetv2}. Compared to the state-of-the-art CNN compression methods, the proposed OD achieves the superior performance. For example, for ResNet-56 on CIFAR-10, the proposed OD with depthwise convolution achieves 6.54$\times$ compression rate and 6.3$\times$ theoretical speedup rate, while only with an increase of 0.36\% error. For ResNet-18 on ImageNet, the proposed OD with depthwise convolution also achieves an increase of 2.46\% error with the highest compression rate of 5.96$\times$ and the theoretical speedup rate of 5.32$\times$, compared to filter pruning methods. Furthermore, the proposed OD can improve the performance of compact networks, such as MobileNet V2 \cite{sandler2018mobilenetv2}.  

We briefly summarize the related work in Section \ref{sec2}. The cheap convolutions we suggest and the proposed online distillation method are described in Section \ref{sec3}.  Section \ref{sec4} presents experimental results. Finally, conclusion is drawn in Section \ref{sec5}.

\section{Related Work}
\label{sec2}
It has shown that CNNs with million-scale parameters typically tend to be heavily over-parameterized \cite{denil2013predicting}, which prohibits their usage on resource-limited environments. Recently, many works have been proposed to compress deep CNNs, which can be further categorized into six groups, \emph{i.e.}, compact network designing, knowledge distillation, network architecture search (NAS), parameter quantization, network pruning and low-rank decomposition. The last four groups are orthogonal to our methods, which can be integrated into our approach to achieve higher performance. For simplicity, we denote them as other orthogonal methods.

\textbf{Compact network designing.} 
As described in Section \ref{sec1}, the key idea of designing a compact model is to replace the loose and over-parametric filters with a compact block to compress and accelerate CNNs. For example, instead of stacking the standard convolutional layers, the inception module in GoogLeNet \cite{szegedy2015going} was proposed to increase the number of branches and depth with much lower computational budget. 
The residual block with bottleneck structure in ResNets \cite{he2016deep} was proposed to achieve remarkable performance, which have been widely-used in a variety of computer vision tasks. 
Huang \emph{et al.} \cite{huang2017densely} proposed DenseNets with the dense block, which encourage feature reuse and substantially reduce the number of parameters.

Recently, one of the most prominent convolution, depthwise separable convolution, has been proposed in \cite{sifre2014rigid}. It applies a depthwise convolution to each input channel, followed by a pointwise convolution over all channels, which has been widely-used in several architectures \cite{chollet2017xception,howard2017mobilenets,sandler2018mobilenetv2}. 
Actually, the depthwise part is an extreme case of group convolution where each partition contains only one channel.
Group convolution was first used in AlexNet \cite{krizhevsky2012imagenet} with two groups for distributing the model over two GPUs to handle the memory issue. 
Subsequently, the works \cite{ioannou2017deep,xie2017aggregated} have also evaluated the better discriminative and generalization ability based on group convolutions than ungrouped counterparts.

Pointwise convolution (\emph{i.e.}, a simple $1\times1$ convolution) is often used after depthwise or group convolution for information recombination across feature channels. However, its expensive computation complexity largely affects the runtime of the state-of-the-art networks \cite{howard2017mobilenets,sandler2018mobilenetv2,huang2018condensenet}. To address the issue, Zhang \emph{et al.} \cite{zhang2018shufflenet} proposed to replace the pointwise convolution with channel shuffle operation, which mixes the channel together to reduce computation cost. To further reduce the practical running time, Ma \emph{et al.} \cite{ma2018shufflenet} design an efficient network, ShuffleNet V2, by following several practical guidelines.

Despite depthwise and group convolution can reduce the computation complexity on the channel dimension, it still cannot reduce the computation overhead on the spatial dimension. To this end, Wu \emph{et al.} \cite{wu2018shift} proposed ShiftNet with shift operations to construct architectures cooperated with pointwise convolution. The shift operation provides spatial information communication by shifting feature maps with zero FLOPs\footnote{FLOPs: The number of Floating-point operations.}. As a generalization of shift operation, an active shift layer \cite{jeon2018constructing} is proposed to formulate the amount of shift as a learnable function with shift parameters. However, each channel with one shifting results in a significant increase in moving memory. Chen \emph{et al.} \cite{chen2019all} proposed a sparse shift layer to eliminate the meaningless shifts.

However, the above compact blocks and cheap convolutions works based on the specific network design. In our method, the cheap convolutions can be generalized to a variety of network architectures by knowledge distillation, especially by online distillation.  

\textbf{Knowledge distillation.}
Knowledge transfer is first exploited to compress model by Buciluǎ \emph{et al.} \cite{bucilua2006model}, in which a simple network can mimic a complicated network without significant loss in performance. 
After that, Hinton \emph{et al.} \cite{hinton2015distilling} proposed dark knowledge (DK) to transfer knowledge from a large pre-trained model to a small student network, which uses an extra supervision provided by the soft final outputs of the pre-trained teacher.
The extra supervision extracted from a pre-trained teacher model is often in form of class posterior probabilities \cite{hinton2015distilling}, feature representations \cite{romero2015fitnets,zagoruyko2017paying,koratana2018lit,lin2018holistic}, distribution of intermediate activations \cite{ahn2019variational,leroux2019training}, or inter-layer flow \cite{yim2017gift}). 
However, these distillation methods require at least two-stage training, including pre-training the teacher network and training the student network with an extra supervision, which is computation expensive during training. 
The more recently proposed deep mutual learning \cite{zhang2018deep} overcomes this limitation by conducting an online distillation in one stage training between two peer student networks. Anil \emph{et al.} \cite{anil2018large} further extended this idea to accelerate the training of large-scale distributed neural networks.
However, these online distillation methods use each student as the opposing teacher, which is not powerful and even limits the efficacy of knowledge discovery.

Actually, the above knowledge distillation methods redesign the new student network, which is thinner or more shallow than the teacher network.
Different from the previous knowledge distillation schemes, we produce a student network by directly replacing the standard convolutions with different cheap convolutions without a complicated redesign.
Moreover, to take the place of peer student as a teacher, we design a new online distillation method to online construct the strong teacher network, in which mutual learning between the teacher and the student model is used to further improve their performance.

\textbf{Other orthogonal methods.}
There are other four kinds of CNN compression methods, \emph{i.e.}, network architecture search \cite{zoph2017neural,zoph2018learning,baker2017designing,real2017large,xie2017genetic,cai2018proxylessnas}, parameter quantization \cite{courbariaux2015binaryconnect,courbariaux2016binarynet,rastegari2016xnor,liang2018fp,jacob2018quantization}, network pruning \cite{han2015learning,han2016deep,li2017pruning,luo2017ThiNet,he2017channel,lin2019toward,lin2019towards,lin2018accelerating,he2018soft} and low-rank decomposition \cite{denton2014exploiting,lin2018holistic,zhang2015efficient,kim2016compression,lebedev2015speeding,lin2016towards}. 
Our method can be integrated with these orthogonal methods to further improve the performance, which are however orthogonal to the core contribution of this paper.
For example, we can prune the student model trained by the proposed OD to be thinner using filter pruning methods \cite{luo2017ThiNet,lin2018accelerating,he2017channel,he2018soft}. 

\section{CNN Compression with Cheap Convolutions}
\label{sec3}
In this section, we first describe the notations. Then, we present several cheap convolutions that may be introduced in place of a standard convolution. Finally, our online distillation methods are introduced to improve the performance of student network with cheap convolutions. 

\subsection{Notations}
\label{ssec3_1}
\begin{figure*}[!t]
\centering
  \subfigure[The standard convolution]{
    \includegraphics[scale = 0.265]{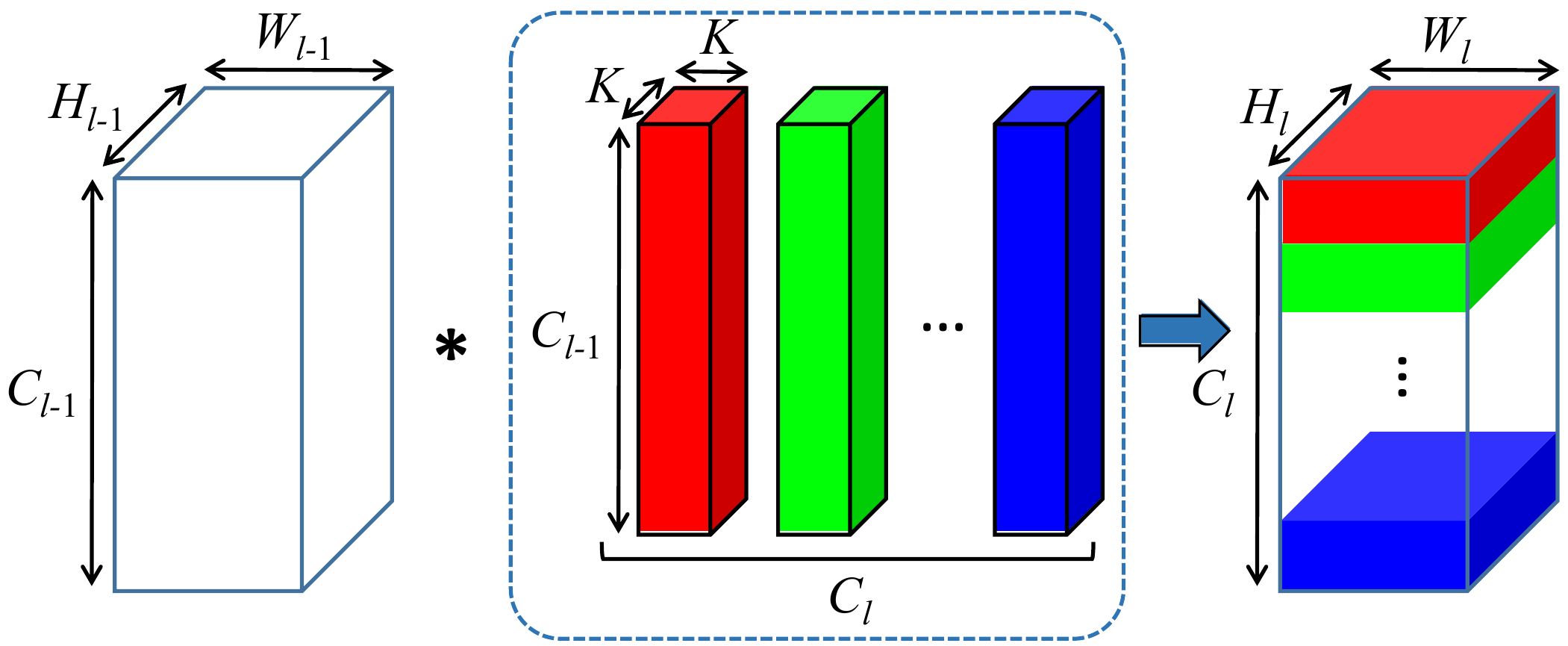}
    \label{fig2_a}
    }
  \subfigure[Group convolution]{
    \includegraphics[scale = 0.265]{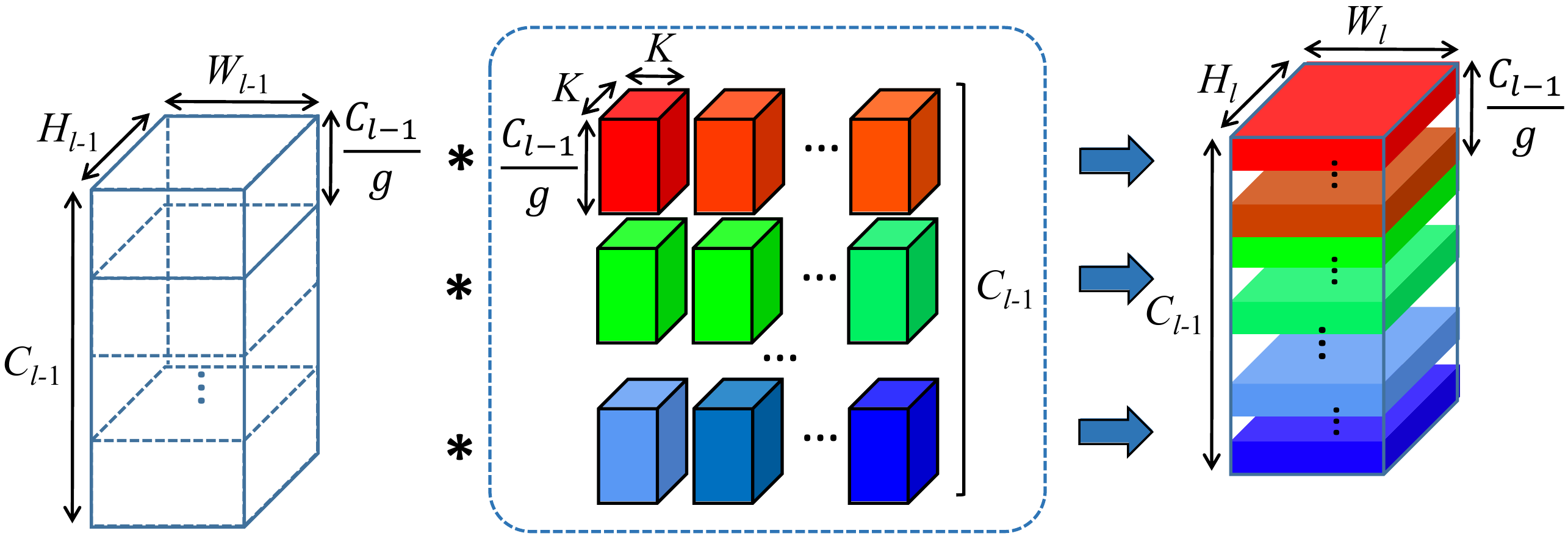}
    \label{fig2_b}
  }
  \subfigure[Depthwise convolution]{
    \includegraphics[scale = 0.3]{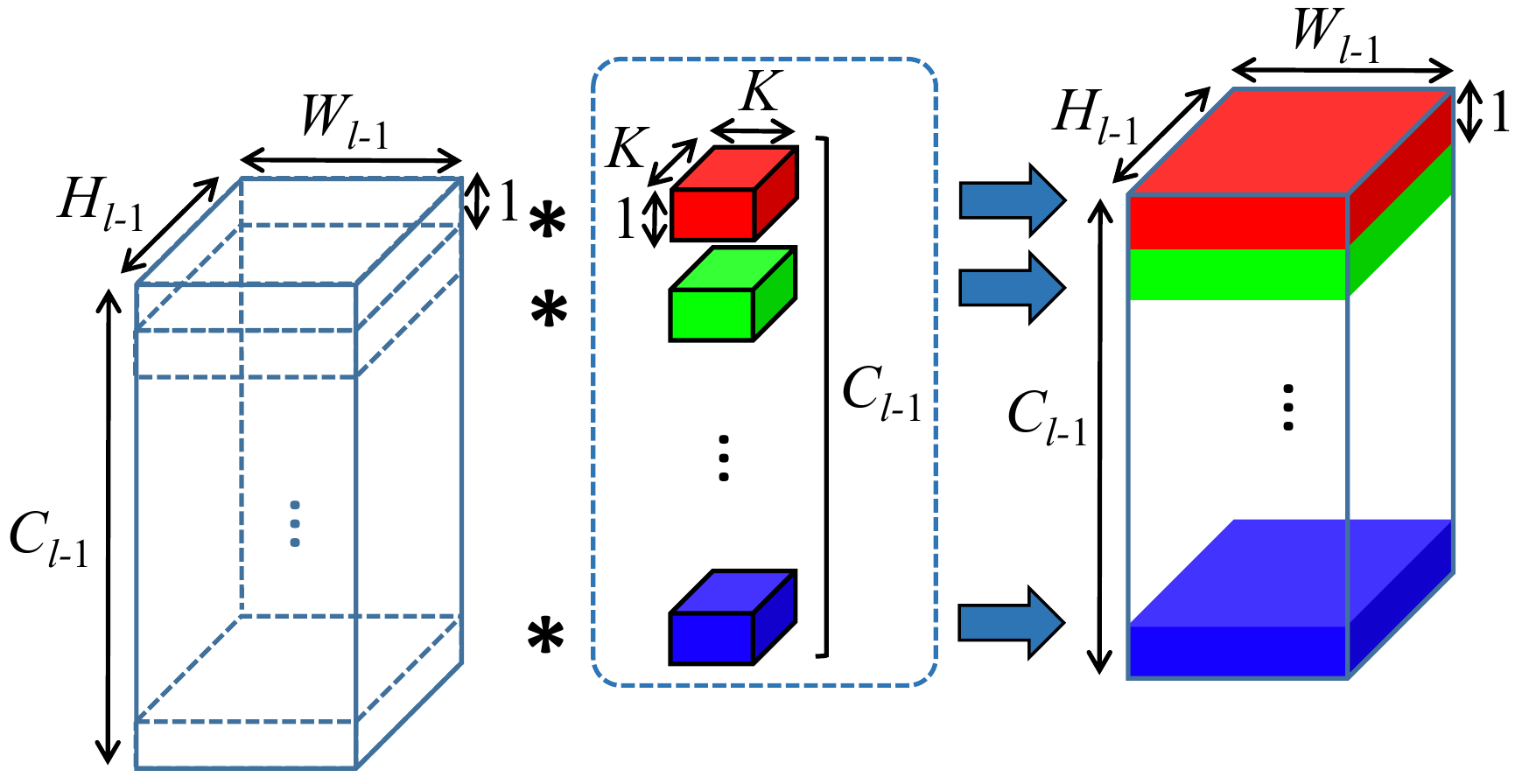}
    \label{fig2_c}
  }
  \subfigure[Shift convolution]{
    \includegraphics[scale = 0.3]{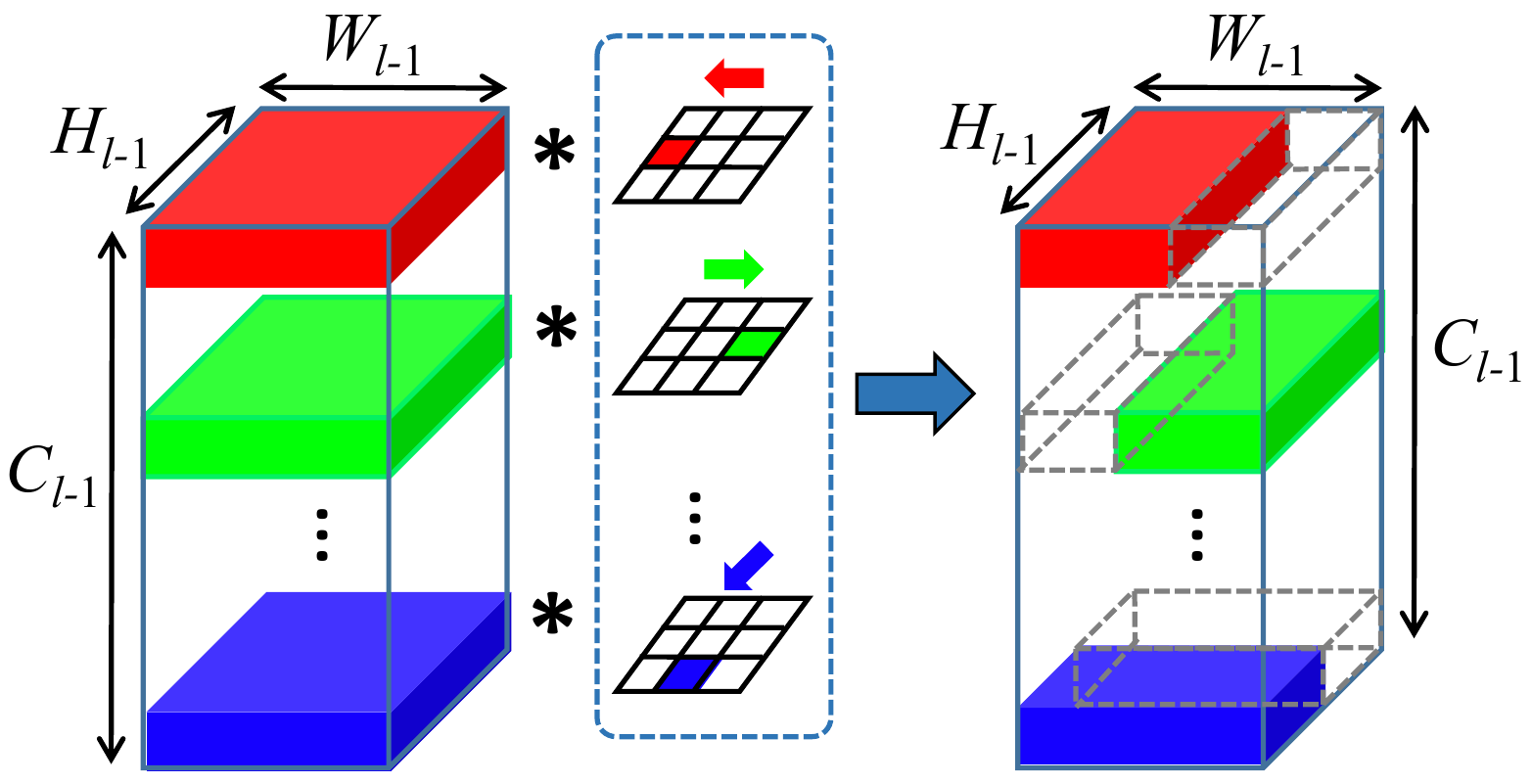}
    \label{fig2_d}
  }
\vspace{-.5em}
\caption{Illustration of several spatial convolutions. (a) the standard convolution, (b) group convolution, (c) depthwise convolution, (d) shift convolution with $3\times3$ shift matrix, in which the color cell denotes 1 at that position, 0 otherwise.}
\label{fig2}
\end{figure*}

Consider a CNN model with $L$ layers in total, which are interlaced with ReLU activation \cite{nair2010rectified}, Batch normalization (BN) \cite{ioffe2015batch} and pooling. Let us denote a set of feature maps in the $l$-th layer by $\mathcal{Z}^l\in\mathbb{R}^{H_l\times W_l\times C_l}$ with the height $H_l$, width $W_l$ and channel size $C_l$.
The feature maps can either be the input of the network $\mathcal{Z}^0$, or the output feature maps $\mathcal{Z}^l, l\in[1,2,\cdots,L]$. 
The convolutional filters/kernels of a spatial convolution in the $l$-th layer is a tensor $\mathcal{K}^l\in\mathbb{R}^{K\times K\times C_{l-1}\times C_l}$. 
For simplicity, we assume the stride is 1 without zero-padding and skip biases.
As shown in Fig. \ref{fig2_a}, the spatial convolution in the $l$-th layer outputs a tensor $\mathcal{Z}^l$, which can be computed as:
\begin{equation}
\label{eq1}
\mathcal{Z}_{h^l,w^l,c_l}^{l} = \sum\limits_{i = 1}^{K}\sum\limits_{j=1}^{K}\sum\limits_{c_{l-1}=1}^{C_{l-1}}\mathcal{K}_{i,j,c_{l-1},c_l}^{l} \mathcal{Z}_{h^{l-1},w^{l-1},c_{l-1}}^{l-1},
\end{equation}
where $i,j$ and $h^{l-1}, w^{l-1}$ index along spatial dimensions at filters and input feature maps in the $l$-th layer, respectively.
$c_{l-1}$ and $c_l$ are the index of input and output channel, respectively. 
The spatial location of the $l$-th outputs is denoted as $h^l=h^{l-1}-i+1$ and $w^l=w^{l-1}-j+1$, respectively.
In the standard convolution, the number of parameters is $C_l\times C_{l-1}\times K^2$ and the computational cost (in FLOPs) is $H_l\times W_l\times C_l\times C_{l-1}\times K^2$.

To further obtain the cross-entropy loss (CE) between the network predictions and the ground-truth label in multi-class classification tasks, we further give a set of training dataset $\mathcal{D}=\{(\mathbf{x}_i,y_i)\}_{i=1}^N$ with $N$ samples, where $\mathbf{x}_i$ and $y_i$ belong to an input and a target output to one of $C$ classes $y_i\in\mathcal{Y}=\{1,2,\cdots,C\}$, respectively.
The network $\mathcal{K}$ (all filters in the entire network) outputs a probabilistic class posterior $p(c|\mathbf{x},\mathcal{K})$ for a sample $\mathbf{x}$ over a class $c$ as:
\begin{equation}
\label{eq2}
p(c|\mathbf{x},\mathcal{K})=\text{softmax}(\mathbf{a})=\frac{exp(\mathbf{a}_c)}{\sum_{i=1}^{C}exp(\mathbf{a}_i)},
\end{equation}
where $\mathbf{a}$ is the logits or the final output before the ``softmax'' operator, which can be computed by:
\begin{equation}
\label{eq3}
\mathbf{a}=FC(\mathcal{Z}^L),
\end{equation} 
where $\mathcal{Z}^L$ is the final output feature maps and $FC(\cdot)$ is the fully-connected layer. 
Therefore, the cross-entropy loss can be computed by:
\begin{equation}
\label{eq4}
\mathcal{L}_{CE}\big(\mathbf{y}, p(c|\mathbf{x},\mathcal{K})\big) = -\sum\limits_{c=1}^{C}\delta_{c,y}\log p(c|\mathbf{x},\mathcal{K}),
\end{equation}
where $\delta_{c,y}$ is the indicator function which returns 1 if $c$ equals to the ground-truth label $y$, and 0 otherwise. For better discussion, the cross-entropy loss between two arbitrary distributions $\mathbf{p}'$ and $\mathbf{q}'$ with the dimension $d$ is formulated as:
\begin{equation}
\label{eq5}
\mathcal{L}_{CE}(\mathbf{p}', \mathbf{q}') = -\sum\limits_{i=1}^{d}\mathbf{p}'_i\log \mathbf{q}'_i.
\end{equation}
Additionally, the Kullback Leibler divergence (KL-divergence) between the distributions $\mathbf{p}'$ and $\mathbf{q}'$ is defined by:
\begin{equation}
\label{eq6}
\mathcal{L}_{KL}(\mathbf{p}', \mathbf{q}') = -\sum\limits_{i=1}^{d}\mathbf{p}'_i\log \frac{\mathbf{p}'_i}{\mathbf{q}'_i}.
\end{equation}

\subsection{Cheap Convolutions}
\label{ssec3_2}
Global Average Pooling (GAP) \cite{lin2013network} is proposed to change the design of fully-connected layers, which reduces the number of fully-connected layers and the network memory overhead. More recent CNN models almost focus on the design of convolutional layers, which are the most time-consuming part and involve the largest number of parameters, such as ResNets \cite{he2016deep}, ResNeXts \cite{xie2017aggregated} and GoogLeNet \cite{szegedy2015going}. 
 In this paper, we present three widely-used cheap convolutions (\emph{i.e.}, group convolution, depthwise convolution and shift convolution) with pointwise convolution in place of a standard convolution to substantially reduce the parameter and computation cost. For better discussion, we consider the convolution in the $l$-th layer.
 
 \textbf{Group Convolution.}
 As shown in Fig. \ref{fig2_b}, group convolution separates the standard convolution into $g$ groups. Generally, it satisfies with $C_{l-1} \% g == 0$, where $\%$ is modulo operator. Then, each group can be computed by:
\begin{footnotesize}
\begin{equation}
\label{eq7}
\mathcal{Z}_{h^l,w^l,c_{l-1}^r}^{l} = \sum_{i = 1}^{K}\sum_{j=1}^{K}\sum_{c_{l-1}=(r-1)C_{l-1}^r+1}^{rC_{l-1}^r}\mathcal{K}_{i,j,c_{l-1},c_{l-1}^r}^{l} \mathcal{Z}_{h^{l-1},w^{l-1},c_{l-1}}^{l-1}, r = 1,2,\cdots, g,
\end{equation}
\end{footnotesize}
\noindent where $C_{l-1}^r = C_{l-1}/g$ is the number of channels in each group and $c_{l-1}^r\in[(r-1)C_{l-1}^r+1,\cdots,rC_{l-1}^r]$ is the index of output channel. 
We obtain a substantial reduction in the number of parameters and the computation cost by only mixing channels within each group. To match the number of output channels $C_l$, pointwise convolution is added after the group convolution. It is a simple implementation that only changes the spatial size of $K\times K$ in the standard convolution to $1\times 1$. Therefore, the total number of parameters and FLOPs in group convolution and pointwise convolution are $C_{l-1}^2\times K^2/g + C_{l-1}\times C_l$ and $H_l\times W_l\times(C_{l-1}^2\times K^2/g + C_{l-1}\times C_l)$, respectively.

\textbf{Depthwise convolution.}
As shown in Fig. \ref{fig2_c}, depthwise convolution \cite{sifre2014rigid} is an extreme case of group convolution, where each group contains only one channel. It means that the number of groups $g$ is setting to the number of input channels $C_{l-1}$. Therefore, the depthwise convolution is computed by simplifying Eq. (\ref{eq7}) as:
\begin{equation}
\label{eq8}
\mathcal{Z}_{h^l,w^l,c_{l-1}}^{l} = \sum\limits_{i = 1}^{K}\sum\limits_{j=1}^{K}\hat{\mathcal{K}}_{i,j,c_{l-1}}^{l} \mathcal{Z}_{h^{l-1},w^{l-1},c_{l-1}}^{l-1},
\end{equation}
where $\hat{\mathcal{K}}\in\mathbb{R}^{K\times K\times C_{l-1}}$ is the depthwise convolution kernels. Then, the following pointwise convolution is added to mix information across channels. Therefore, the total number of parameters and FLOPs in depthwise convolution and pointwise convolution are $C_{l-1}\times K^2 + C_{l-1}\times C_l$ and $H_l\times W_l\times(C_{l-1}\times K^2 + C_{l-1}\times C_l)$.

\textbf{Shift convolution.}
As illustrated in Fig. \ref{fig2_d}, shift convolution \cite{wu2018shift} is a special case of depthwise convolution by replacing depthwise convolution kernels with shift kernels. Specifically, it can be computed logically by:
\begin{equation}
\label{eq9}
\mathcal{Z}_{h^l,w^l,c_{l-1}}^{l} = \sum\limits_{i = 1}^{K}\sum\limits_{j=1}^{K}\bar{\mathcal{K}}_{i,j,c_{l-1}}^{l} \mathcal{Z}_{h^{l-1},w^{l-1},c_{l-1}}^{l-1},
\end{equation}
where $\bar{\mathcal{K}}\in\mathbb{R}^{K\times K\times C_{l-1}}$ is the shift kernels such that
\begin{equation}
\label{eq10}
\bar{\mathcal{K}}_{i,j,c_{l-1}}^{l}=\left \{ \begin{aligned}
1, & \quad i=i_{c_{l-1}} \, \text{and} \, j=j_{c_{l-1}},\\
0, & \quad otherwise.\\
\end{aligned}
\right .
\end{equation}
Here $i_{c_{l-1}},j_{c_{l-1}}$ are channel-dependent indices that assign one of the value in $\bar{\mathcal{K}}_{:,:,c_{l-1}}^{l}\in\mathbb{R}^{K\times K}$ to be 1 and the rest to 0. 
In fact, shift kernels can be implemented by channel permutation operator such that the shift operation itself does not require paramenters or FLOPs. 
We also fuse the shift convolution with the following pointwise convolution, which fetches data from the shifted address in memory and mixes information across channels.
Therefore, the total number of parameters and FLOPs in shift convolution and pointwise convolution are $C_{l-1}\times C_l$ and $H_l\times W_l\times C_{l-1}\times C_l$.

\begin{table*}[t]
\tiny
\begin{center}
\caption{Comparison of several convolutions in parameters and FLOPs at the $l$-th layer. CR and SR represent compression rate and speedup rate based on $C_{l-1}=C_l$, respectively.}
\label{tab2}
\begin{tabular}{ccccc}
\hline
Convolution  & \#Param. & FLOPs & CR($\times$) & SR($\times$) \\
\hline
Standard & $C_l\times C_{l-1}\times K^2$ & $H_l\times W_l\times C_l\times C_{l-1}\times K^2$ & 1 & 1 \\ \hline
Group + $1\times1$ & $C_{l-1}^2\times K^2/g + C_{l-1}\times C_l$ & $H_l\times W_l\times(C_{l-1}^2\times K^2/g + C_{l-1}\times C_l)$ & $\frac{K^2g}{K^2+g}$ & $\frac{K^2g}{K^2+g}$ \\ \hline
Depthwise + $1\times1$ & $C_{l-1}\times K^2 + C_{l-1}\times C_l$ & $H_l\times W_l\times(C_{l-1}\times K^2 + C_{l-1}\times C_l)$ & $\frac{K^2C_l}{K^2+C_l}$ & $\frac{K^2C_l}{K^2+C_l}$\\ \hline
Shift + $1\times1$ & $C_{l-1}\times C_l$ & $H_l\times W_l\times C_{l-1}\times C_l$ & $K^2$ & $K^2$ \\
\hline
\end{tabular}
\end{center}
\end{table*}

\textbf{Comparison of cheap convolutions in parameters and FLOPs.} 
We further compare the standard convolution with the three cheap convolution in parameters and FLOPs. 
The results are summarized in Table \ref{tab2}. 
For the same number of input and output channel (\emph{i.e.}, $C_{l-1}=C_l$), the group convolution with the following pointwise convolution achieves $K^2g/(K^2+g)\times$ compression and speedup rates, compared to the standard convolution. Assuming the $3\times3$ spatial convolution with 4 groups, the compression and speedup rate will attained $2.76\times$. Their values are increasing with an increase of the number of groups. The depthwise convolution is an extreme case of group convolution by separating each group with one channel. Both compression and speedup rates in depthwise convolution with pointwise convolution attain to $K^2C_l/(K^2+C_l)\times$, which are much higher than group convolution. 
Instead of the depthwise convolution with $K\times K$ 32-bit float kernels, the shift convolution employs the free shift operations with the special binary values, in which only one of the values in each 2D kernel assigns to be 1 and the rest to be 0. Therefore, the shift convolution followed by pointwise convolution achieves $K^2\times$ compression and speedup rates, which are extremely higher than depthwise convolution.

\subsection{Online Distillation}
\label{ssec3_3}
By replacing the standard convolution with several cheap convolutions, a variety of CNN models can be compressed and accelerated substantially. It is simple and straightforward to train the compressed models with cheap convolutions from scratch to improve the accuracy. By doing this, it however leads to a limited improvement on accuracy, which is due to a limited knowledge used only by the ground-truth labels. 

Alternatively, knowledge distillation \cite{hinton2015distilling,zagoruyko2017paying} is becoming a promising solution, which aims to transfer more knowledge from a teacher network to a student network to boost the accuracy of the student network.
For this paper, we first review two different distillation methods for learning a smaller student network from a large, pre-trained teacher network: dark knowledge (DK) \cite{hinton2015distilling} and attention transfer (AT) \cite{zagoruyko2017paying}.
In that case, we can select a CNN model with the standard convolution as a teacher, while a model with cheap convolution by keeping the teacher's architecture is regarded as a student. 
Then, we propose our online distillation (OD) method to replace the pre-trained teacher network by constructing online from the multiple student networks with the same architecture, and train both teacher and student networks in a one-shot manner.

\textbf{Dark Knowledge.}
Let $t$ and $s$ be a teacher network and a student network with the final output features $\mathcal{Z}_{(t)}^{L}$ and $\mathcal{Z}_{(s)}^{L}$, respectively. $\mathbf{a}_{(t)}$ and $\mathbf{a}_{(s)}$ are the logits of teacher and student networks, which can be computed respectively by:
\begin{equation}
\label{eq11}
\mathbf{a}_{(t)}=FC(\mathcal{Z}_{(t)}^{L}), \quad \mathbf{a}_{(s)}=FC(\mathcal{Z}_{(s)}^{L}).
\end{equation}
Given an input image $\mathbf{x}$, the probabilistic class posterior of teacher and student network $p(c|\mathbf{x}, \mathcal{K}_{(t)})$ and $p(c|\mathbf{x}, \mathcal{K}_{(s)})$ over a class $c$ can be computed respectively as:
\begin{equation}
\label{eq12}
p(c|\mathbf{x}, \mathcal{K}_{(t)})=\text{softmax}(\mathbf{a}_{(t)}), \quad p(c|\mathbf{x}, \mathcal{K}_{(s)})=\text{softmax}(\mathbf{a}_{(s)}),
\end{equation}
where $\mathcal{K}_{(t)}$ and $\mathcal{K}_{(s)}$ are the parameters in the teacher and student networks.
To perform dark knowledge, we train the student network to minimize the following loss function:
\begin{equation}
\small
\label{eq13}
\mathcal{L}_{DK} = \mathcal{L}_{CE}\big(\mathbf{y},p(c|\mathbf{x}, \mathcal{K}_{(s)})\big)+\tau^2\mathcal{L}_{CE}\big(\text{softmax}(\mathbf{a}_{(t)}/\tau), \text{softmax}(\mathbf{a}_{(s)}/\tau)\big),
\end{equation}
where $\tau$ is a temperature parameter to soften the distribution of predictions. Different from fine-tuning just by class labels, dark knowledge is to learn the student network from the soft outputs of teacher network and ground-truth labels. 

\textbf{Attention transfer.}
Instead of the loss on the soft outputs between teacher and student networks in Eq. (\ref{eq13}), attention transfer constructs the losses between teacher and student attention maps to learn the student model.
Let $\mathcal{I}$ denote the indices of all teacher-student activation layer pairs for which needs to transfer attention maps. For each $l\in\mathcal{I}$, we first collect the $j$-th feature map in the $l$-th layer of teacher and student networks into the vector $\mathbf{z}_{(t)j}^{l}$ and $\mathbf{z}_{(s)j}^{l}$, respectively. After that, we collect the $\mathbf{z}_{(t)j}^{l}$ and $\mathbf{z}_{(s)j}^{l}$ for all $j$, which are denoted as $\mathbf{Z}_{(t)}^{l}$ and $\mathbf{Z}_{(s)}^{l}$, respectively. Then the total loss of attention transfer can be formulated by:
\begin{equation}
\small
\label{eq14}
\mathcal{L}_{AT} = \mathcal{L}_{CE}\big(\mathbf{y},p(c|\mathbf{x}, \mathcal{K}_{(s)})\big)+\beta\sum\limits_{l\in\mathcal{I}}\big\|\frac{f(\mathbf{Z}_{(t)}^{l})}{\|f(\mathbf{Z}_{(t)}^{l}\|_2}-\frac{f(\mathbf{Z}_{(s)}^{l})}{\|f(\mathbf{Z}_{(s)}^{l}\|_2}\big\|_2,
\end{equation}
where $\beta$ is a hyperparameter, and $f(\mathbf{Z}^l) = \frac{1}{C_l}\sum_{j=1}^{C_l}{\mathbf{z}_{j}^{l}}^2$. Note that the attention transfer uses the $\ell_2$-normalized attention maps, which are important for the success of the student training.

\textbf{Online distillation.}
Different from DK and AT by the pre-trained teacher network, the proposed online distillation method construct the pre-trained teacher network online. 
As presented in Fig. \ref{fig1_b}, the strong teacher network is formed by concatenating all the features from the last convolutional outputs of multiple branch networks\footnote{A branch is regarded as one student network.} with the same architectures and then adding the following BN, ReLU and GAP, one fully-connected layer.
From the respect of network architecture, it looks like the siamese networks \cite{koch2015siamese,zagoruyko2015learning}, but they are totally different in the following aspects: 
\begin{itemize}
\item[1.] \emph{Input.} The inputs in our online distillation are a mini-batch of images, while the inputs in siamese networks are two/triple pair images. 
\item[2.] \emph{Output.} The number of output nodes in our online distillation is the number of classes, while the corresponding output number in siamese networks is only one.
\item[3.] \emph{Function.} Online distillation aims to improve the discriminative ability of the student network by transferring the knowledge from the strong teacher. In contrary, learning the similarity of pair-image is the main target in the siamese networks.
\end{itemize}

Let $\mathcal{Z}_{(s)}^{L}, s\in[1,2,\cdots,m]$ denote the last convolutional outputs in the $s$-th student network, where $m$ is the number of student networks.
By doing this, the logits $\mathbf{a}_{(t)}$ in the strong teacher network can be formulated as:
\begin{equation}
\label{eq15}
\mathbf{a}_{(t)}=FC\bigg(G\Big(\big[\mathcal{Z}_{(1)}^{L}, \mathcal{Z}_{(2)}^{L}, \cdots, \mathcal{Z}_{(m)}^{L}\big]\Big),\theta\bigg),
\end{equation}
where $G$ is the transformation function including BN, ReLU and GAP. $\theta$ is the parameter in the fully-connected layer of teacher network.
The design of multi-branch has two merits for model training: (1) A strong teacher network is created online, which can be simultaneously trained with several student networks in an end-to-end manner. (2) A complex asynchronous update between multiple networks can be avoided for fast training. 

After obtaining the strong teacher network, we distill the knowledge from it to all the student networks. Followed by dark knowledge, the soft predictions between the teacher and each student are aligned to improve their accuracy using the KL-divergence. Therefore, the overall loss function can be formulated as:
\begin{equation}
\scriptsize
\label{eq16}
\mathcal{L}_{OD} = \sum\limits_{s=1}^{m}\mathcal{L}_{CE}\big(\mathbf{y},p(c|\mathbf{x}, \mathcal{K}_{(s)})\big)+\mathcal{L}_{CE}\big(\mathbf{y},p(c|\mathbf{x}, \mathcal{K}_{(t)})\big)+\tau^2\sum\limits_{s=1}^{m}\mathcal{L}_{KL}\big(\text{softmax}(\mathbf{a}_{(t)}/\tau), \text{softmax}(\mathbf{a}_{(s)}/\tau)\big),
\end{equation}
where $\mathcal{K}_{(s)}$ and $\mathcal{K}_{(t)}=\{\cup_{s=1}^{m}\mathcal{K}_{(s)},\theta\}$ are the parameters of each student network and teacher network, respectively. $\mathbf{a}_{(s)}$ is the logits of the $s$-th student network.

\section{Experiments}
\label{sec4}
In this section, we train and evaluate a number of student networks with cheap convolutions using the proposed online distillation, as well as dark knowledge and attention transfer. Comprehensive experiments are conducted on three widely-used datasets, CIFAR-10 \cite{krizhevsky2009learning}, CIFAR-100 \cite{krizhevsky2009learning} and ImageNets \cite{russakovsky2015imagenet}.  
CIFAR-10/100 consists of 60,000 $32\times32$ color images (50,000 training images and 10,000 test images) from 10/100 classes, with 6,000/600 images per class. 
In our implementation, we randomly select 10\% training images (\emph{i.e.}, 500/50 images per class for CIFAR-10/100) as validation dataset.
ImageNet ILSVRC 2012 contains 1.28M training images and 50,000 validation images from 1,000 classes. 
For ImageNet, we rescale the training images to the size of $256\times256$, with a $224\times224$ crop randomly sampled from each image and its horizontal flip \cite{krizhevsky2012imagenet}. The performance of student models are tested on the validation set using the single-view testing with only the central patch. We also evaluated the performance on different kinds of network architectures, ResNets \cite{he2016deep}, DenseNets \cite{huang2017densely}, WideResNets (WRN)\cite{zagoruyko2016wide} and MobileNets \cite{sandler2018mobilenetv2}. 

We train the student networks by PyTorch \cite{paszke2017automatic} and run them on four NVIDIA GTX 1080Ti GPUs. The weight decay is set to 0.0002 and momentum is set to 0.9. For CIFAR-10 and CIFAR-100, we use the same training parameters, in which the mini-batch size is set to 128 for 300 epochs, and the  learning rate is initialized by 0.1 with a constant dropping factor of 10 throughout 50\% and 75\% of the total number of epochs.
For ImageNet, we set the same weight decay, momentum and initialized learning rate to CIFAR-10/100. The mini-batch size and the total epoch are set to 64 and 100, respectively. The learning rate is step dropping in a factor of 10 after 30 epochs. For $\beta$ in Eq. (\ref{eq14}), we fix it to be 1. 
For the number of branches $m$ and temperature $\tau$, their setups will be discussed subsequently in the following ablation study.
For online distillation (OD), the student model with the best performance is selected to test its final performance in test dataset\footnote{For ImageNet, the performance of the student model with the best in training dataset is selected to test its final performance in validation dataset.}.
For attention transfer (AT) \cite{zagoruyko2017paying}, we use the knowledge from attention maps at the final layer of each stage, but not including the knowledge from the probabilistic class posterior of teacher network.
The training parameters in dark knowledge (DK) \cite{hinton2015distilling} and attention transfer are same to OD.

\subsection{Ablation Study}
\label{ssec4_1}
In this section, we explore the effect of the number of student branches $m$ and temperature $\tau$ on the proposed online distillation, as well as different knowledge distillation methods. We select the ResNet-56 on CIFAR-10 as a given architecture or baseline, and use the shift convolution as a cheap convolution. For ResNet-56, it consists of 27 residual blocks, which achieves the error of 6.20\% with the number of parameters 0.85M and FLOPs 126.81M. Each block has two convolutional layers with a standard kernel size of $3\times 3$. Instead of the standard convolution, shift convolution and pointwise convolution are applied into each block, termed \emph{Shift-Point-Shift-Point} or \emph{SPSP} module, to reduce the computation and memory cost.

\begin{figure*}[t]
\centering
  \subfigure[Different temperatures]{
    \includegraphics[scale = 0.325]{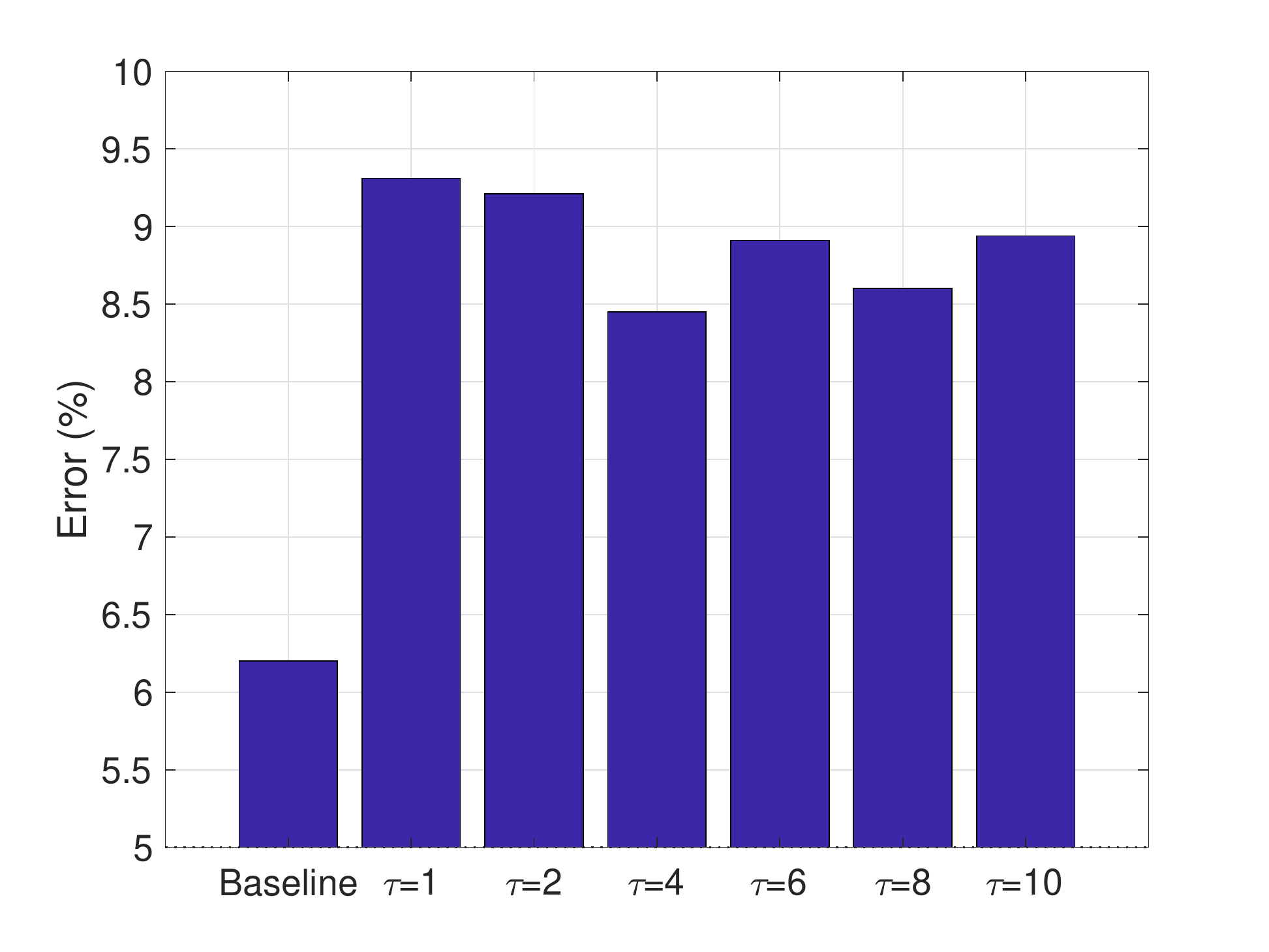}
    \label{fig3_a}
    }
  \subfigure[Model complexity]{
    \includegraphics[scale = 0.325]{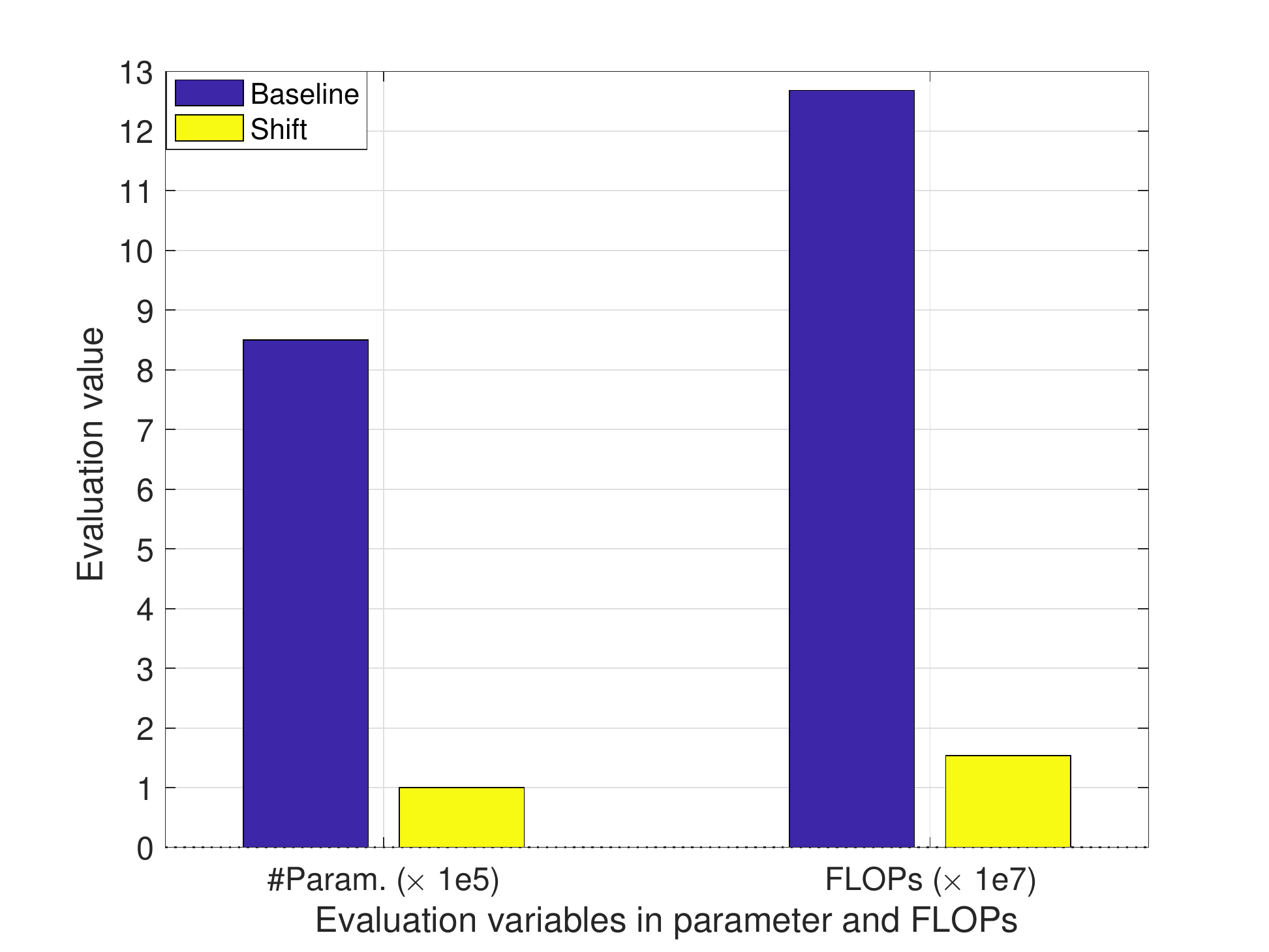}
    \label{fig3_b}
  }
\vspace{-1em}
\caption{Results of online distillation with different temperatures.}
\label{fig3}
\end{figure*}

\textbf{Effect on the temperature.}
To evaluate the effect of the temperatures on online distillation, we fix the branch number $m$ to 4 and set the same training parameters, such as learning rate, epoch number and learning strategy. 
We train OD with a variety of temperatures by 1, 2, 4, 6, 8 and 10. The results are summarized in Fig. \ref{fig3}. The effect of temperature on the performance is shown in Fig. \ref{fig3_a}. We can see that temperature $\tau$ is set to 4, which achieves the best performance compared to other values. This is due to that over-high temperature forces the softened prediction distributions between teacher and each student network to be similar, which reduces the flow of knowledge. 
Instead, a low temperature also affects the performance, \emph{e.g.}, temperature $\tau$ is set to 1. This finding is actually consistent with the work \cite{hinton2015distilling}. Therefore, to simplify experiments, we set temperature to 4 in all of the following experiments. 
As shown in Fig. \ref{fig3_b}, we also found that replacing the standard convolution with the shift and pointwise convolution results in a significant reduction on the number of parameters and FLOPs, but also with a higher increase of error, compared to the baseline. 

\begin{figure*}[!t]
\begin{center}
\includegraphics[scale=0.43]{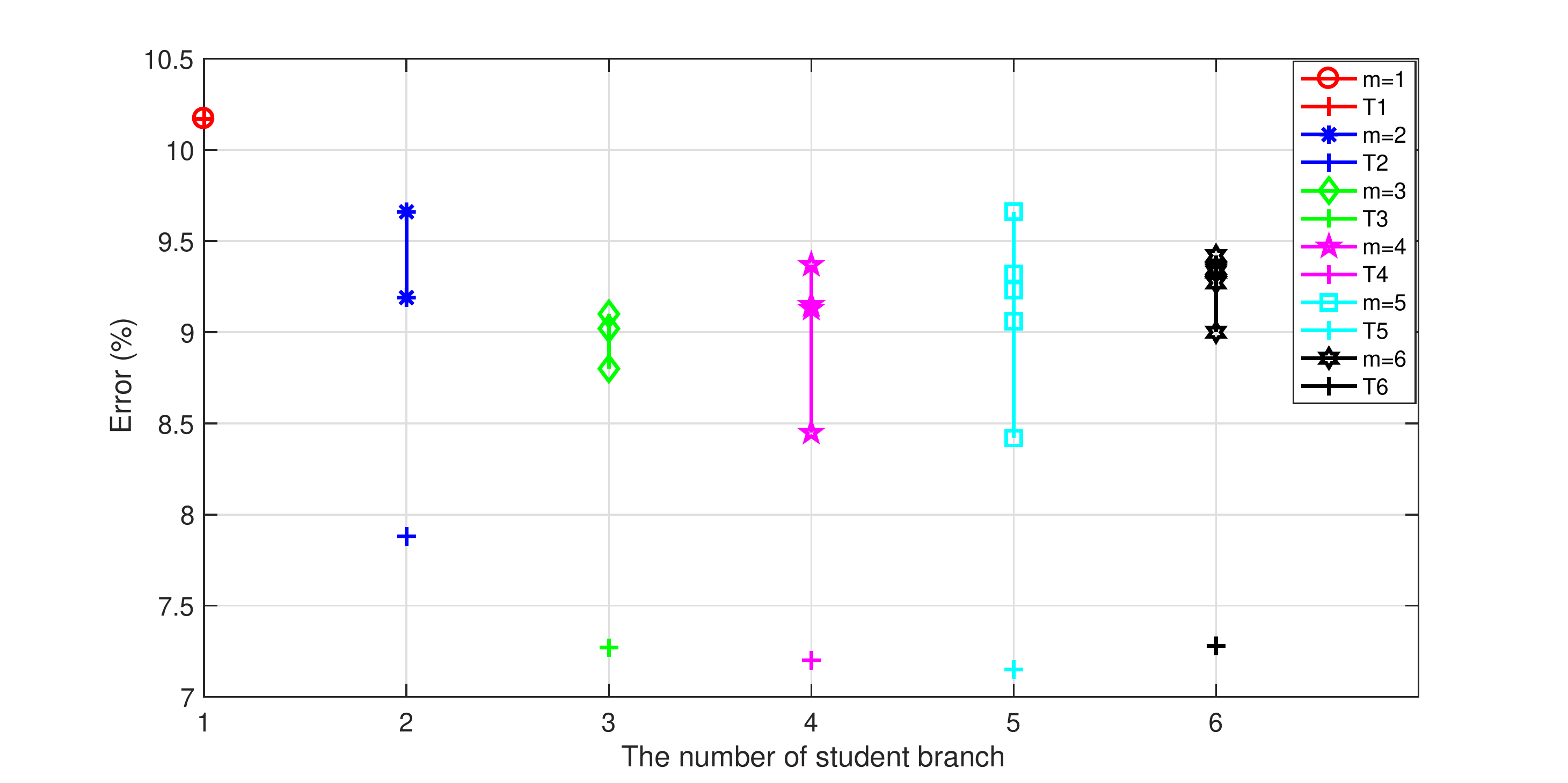}
\end{center}
\caption{Results of the different branch number. $m=i$ and T$i$ represent the number of student branches $i$ and the teacher model by aggregating the number of $i$ students, respectively.}
\label{fig4}
\end{figure*}

\textbf{Effect on the number of student branches.}
To evaluate the effect of the student branch number, we fix the temperature to 4 and only change the branch number from 1 to 6. When the branch number is set to 1, it means that the student network is the same to the teacher one. 
As shown in Fig. \ref{fig4}, with an increase of branch number, the error of teacher model decrease gradually, except the T6 with the number of 6 branches. We conjecture this is due to the over-parameters in T6 instead affects the discriminative ability of teacher model. Correspondingly, the student model also achieves the consistent results. 
By setting $m$ to 4 and 5, we can obtain the best student models with almost the same performance. For example, compared to branch number of 1, 2, 3 and 6, OD with branch number of 4 and 5 achieve significantly higher performance with classification error of 8.45\% and 8.42\%, respectively. 
Actually, with an increase of branch number, the training speed will certainly slow down with an increasing GPU RAM overhead.
Furthermore, to maximize the parallel computing speed in the heterogeneous computing platform, we set the number of branches to 4 in subsequent experiments, which matches the number of our GPUs.  
 Therefore, we report the result by fixing both the number of branches and temperature to 4 unless otherwise stated.

\begin{table*}[t]
\footnotesize
\begin{center}
\caption{Comparison of different knowledge distillation methods on ResNet-56.}
\label{tab3}
\begin{tabular}{ccccc}
\hline
Method  & Convolution & Error \% & \#Params (M) & FLOPs (M) \\
\hline
ResNet-56 & Standard & 6.20 & 0.85 & 126.81 \\ 
\hline
Training from scratch & Shift & 10.17 & 0.10 & 15.39 \\
DK \cite{hinton2015distilling} & Shift & 9.07 & 0.10 & 15.39 \\
AT \cite{zagoruyko2017paying} & Shift & 8.84 & 0.10 & 15.39 \\ 
OD & Shift & \textbf{8.45} & 0.10 & 15.39 \\ 
\hline \hline
Training from scratch & Depthwise & 9.55 & 0.13 & 20.14 \\
DK \cite{hinton2015distilling} & Depthwise & 6.75 & 0.13 & 20.14 \\
AT \cite{zagoruyko2017paying} & Depthwise & 7.94 & 0.13 & 20.14 \\ 
OD & Depthwise & \textbf{6.58} & 0.13 & 20.14 \\ 
\hline \hline
Training from scratch & Group-16 & 8.58 & 0.16 & 23.35 \\
DK \cite{hinton2015distilling} & Group-16 & 6.90 & 0.16 & 23.35 \\
AT \cite{zagoruyko2017paying} & Group-16 & 7.14 & 0.16 & 23.35 \\ 
OD & Group-16 & \textbf{6.12} & 0.16 & 23.35 \\ 
\hline \hline
Training from scratch & Group-8 & 8.37 & 0.21 & 31.31 \\
DK \cite{hinton2015distilling} & Group-8 & 6.94 & 0.21 & 31.31 \\
AT \cite{zagoruyko2017paying} & Group-8 & 6.54 & 0.21 & 31.31 \\ 
OD & Group-8 & \textbf{6.12} & 0.21 & 31.31 \\
\hline \hline
Training from scratch & Group-4 & 8.33 & 0.31 & 47.24 \\
DK \cite{hinton2015distilling} & Group-4 & 6.87 & 0.31 & 47.24 \\
AT \cite{zagoruyko2017paying} & Group-4 & 6.25 & 0.31 & 47.24\\ 
OD & Group-4 & \textbf{6.22} & 0.31 & 47.24 \\
\hline \hline
Training from scratch & Group-2 & 7.88 & 0.52 & 79.09 \\
DK \cite{hinton2015distilling} & Group-2 & 6.52 & 0.52 & 79.09 \\
AT \cite{zagoruyko2017paying} & Group-2 & \textbf{6.38} & 0.52 & 79.09 \\ 
OD & Group-2 & 6.45 & 0.52 & 79.09 \\
\hline 
\end{tabular}
\end{center}
\end{table*}

\subsection{Comparison with the State-of-the-art}
In this section, we first make a comparison of the different distillation methods on the same cheap convolution, and then the different cheap convolutions on the same knowledge distillation. Finally, we compare our method to the state-of-the-art CNN compression methods.

\subsubsection{OD vs. DK and AT}
As described in Section \ref{ssec3_3}, DK \cite{hinton2015distilling} and AT \cite{zagoruyko2017paying} require a pre-trained teacher model.
For a fair comparison, we choose the standard ResNet-56 as their teachers. The results of different knowledge distillation methods are summarized in Table \ref{tab3}. 
With the same cheap convolution, the proposed OD achieves the best performance, compared to DK and AT. For example, OD achieves the lowest error of 8.45\% (\emph{vs.} 9.07\% in DK and 8.84\% in AT), when the standard convolution is replaced with shift and pointwise convolution. We also found that the compressed model with group convolution trained by OD even achieves better performance than the standard ResNet-56. For example, by setting the number of groups $g$ to 16, OD achieves a lower error of 6.12\% with a smaller amount of parameters 0.16M and FLOPs 23.35M, compared to 6.20\% with in ResNet-56. 
In addition, by increasing $g$ from 2 to 16 in OD, the number of parameters and FLOPs are decreasing gradually, while the performance is improved steadily. We conjecture this is due to the better generalization ability by ensembling the students with a smaller amount of parameters.
Note that AT achieves the best performance with a  error of 6.38\% by setting $g$ to 2, compared to DK and OD. 

\begin{table*}[t]
\scriptsize
\begin{center}
\caption{Comparison of different cheap convolutions using OD.}
\label{tab4}
\begin{tabular}{ccccc}
\hline
Model & Convolution & Error \% & \#Params (M) & FLOPs (M) \\
\hline
\multirow{7}{*}{ResNet-56 on CIFAR-10 \cite{he2016deep}} & Standard & 6.20 & 0.85 & 126.81 \\ 
& Shift & 8.45 & \textbf{0.10} & \textbf{15.39}\\
& Depthwise & 6.58 & 0.13 & 20.14\\
& Group-16 & \textbf{6.12} & 0.16 & 23.35\\
& Group-8 & \textbf{6.12} & 0.21 & 31.31\\
& Group-4 & 6.22 & 0.31 & 47.24\\
& Group-2 & 6.45 & 0.52 & 79.09\\
\hline \hline
\multirow{7}{*}{ResNet-56 on CIFAR-100 \cite{he2016deep}} & Standard & 27.84 & 0.86 & 126.82 \\ 
& Shift & 32.19 & \textbf{0.11} & \textbf{15.39}\\
& Depthwise & 30.46 & 0.13 & 20.15\\
& Group-16 & 29.92 & 0.16 & 23.35\\
& Group-8 & \textbf{29.23} & 0.21 & 31.32\\
& Group-4 & 30.14 & 0.32 & 47.24\\
& Group-2 & 29.54 & 0.53 & 79.09\\
\hline \hline
\multirow{4}{*}{DenseNet-40-12 on CIFAR-10 \cite{huang2017densely}} & Standard & 5.19 & 1.06 & 282.92 \\ 
& Shift & 6.41 & \textbf{0.25} & \textbf{79.87}\\
& Depthwise & 5.79 & 0.33 & 98.91\\
& Group-12 & \textbf{5.24} & 2.15 & 383.74\\
\hline 
\end{tabular}
\end{center}
\end{table*}

\subsubsection{Comparison of different cheap convolutions}
We further evaluate the performance of different cheap convolutions based on OD. For DenseNet-40-12 \cite{huang2016deep}, it has 40 layers with the growth rate of 12, which means that the number of output channels in each dense block is set to 12.
As shown in Table \ref{tab4}, the networks with shift convolution require a smaller amount of parameters and FLOPs, which however leads to a largest increase on the classification error, compared to depthwise and group convolutions.
For group convolution, by effectively selecting the number of group $g$, the compressed networks can achieve the performance beyond the standard network. For example, the compressed ResNet-56 on CIFAR-10 with the group number of 16 achieves 6.12\% error against the original network with 6.20\% error.
It also indicates that OD performs mutual learning between a strong online teacher and individual students, which improves the performance of student models.
We also found that the networks with depthwise convolution achieve the best trade-off between error and the number of parameters. For example, DenseNet-40-12 with depthwise convolution achieves only an increase of 0.6\% error and results in $3.21\times$ compression rate, compared to the original DenseNet-40-12.
Note that applying group convolution with $g=12$ to DenseNet-40-12 significantly increases the number of parameter and FLOPs instead. This is due to the number of input channels is significantly larger than the number of output channels (\emph{i.e.}, 12) with the dense connectivity pattern, and the proposed group convolution is implemented by setting the number of output channels to be equal to the number of input channels.
Therefore, shift and depthwise convolutions are more suitable for compressing DenseNets, compared to group convolution.

\begin{table*}[!t]
\footnotesize
\begin{center}
\caption{Results of compressing different models on CIFAR-10. OD-Shift/Depthwise/Group represents the combination of OD and Shift/Depthwise/Group convolution. OD-S$i$ and OD-T refer to the $i$-th student and the teacher in OD, respectively.}
\label{tab5}
\begin{tabular}{ccccc}
\hline
Model & Method & Error \% & \#Params (M) & FLOPs (M) \\
\hline
\multirow{14}{*}{ResNet-56} & Baseline & 6.20 & 0.85 & 126.81 \\ 
& He \emph{et al.} \cite{he2017channel} & 8.20 & - & 62 \\
& L1 \cite{li2017pruning} & 6.94 & 0.73 & 90.9 \\
& NISP \cite{yu2018nisp} & 6.99 & 0.49 & 81 \\
& AMC \cite{he2018amc} & 8.1 & - & 63.4 \\
& SFP \cite{he2018soft} & 6.65 & - & 59.4 \\
& C-SGD \cite{ding2019centripetal} & 6.56 & - & 49.6 \\
& PFGM \cite{he2019filter} & 6.51 & - & 59.4 \\
& GAL-0.6 \cite{lin2019towards} & 6.62 & 0.75 & 78.30 \\
& GAL-0.8 \cite{lin2019towards} & 8.42 & 0.29 & 49.99 \\
& ResNet-56-P2 \cite{singh2019hetconv} & 6.40 & - & 71.45 \\
& ResNet-56-P4 \cite{singh2019hetconv} & 6.71 & - & 42.28 \\
& OD-Depthwise & 6.58 & \textbf{0.13} & \textbf{20.14} \\
& OD-Group-16 & \textbf{6.12} & 0.16 & 23.35\\
\hline \hline
\multirow{5}{*}{WRN-40-1} & Baseline & 6.46 & 0.56 & 83.02 \\ 
& WRN-40-2-Depthwise \cite{crowley2018moonshine} & \textbf{6.5} & 0.29 & -\\
& OD-Shift & 8.01 & \textbf{0.07} & \textbf{10.67} \\
& OD-Depthwise & 6.96 & 0.08 & 13.87 \\
& OD-Group-16 & 6.71 & 0.10 & 15.98 \\
\hline \hline
\multirow{8}{*}{DenseNet-40-12} & Baseline & 5.19 & 1.06 & 282.92 \\ 
& Liu \emph{et al.}-40\% \cite{liu2017learning} & \textbf{5.19} & 0.66 & 190 \\
& Liu \emph{et al.}-70\% \cite{liu2017learning} & 5.65 & 0.35 & 120 \\
& GAL-0.05 \cite{lin2019towards} & 5.50 & 0.45 & 128.11 \\
& GAL-0.1 \cite{lin2019towards} & 6.77 & 0.26 & 80.89 \\
& C-SGD \cite{ding2019centripetal} & 5.56 & - & 113.02 \\
& OD-Shift & 6.41 & \textbf{0.25} & \textbf{79.87}\\
& OD-Depthwise & 5.79 & 0.33 & 98.91\\
\hline  \hline
\multirow{6}{*}{MobileNet V2 \cite{sandler2018mobilenetv2}} & Baseline & 6.72 & 2.26 & 91.15 \\ 
& OD-S1 & 6.56 & 2.26 & 91.15\\
& OD-S2 & 6.45 & 2.26 & 91.15\\ 
& OD-S3 & 6.38 & 2.26 & 91.15\\
& OD-S4 & \textbf{5.99} & 2.26 & 91.15\\
& OD-T & 5.18 & - & - \\
\hline
\end{tabular}
\end{center}
\end{table*}

\subsubsection{CIFAR-10}
We evaluate the performance of the proposed OD with several cheap convolutions on CIFAR-10 and compare them with several state-of-the-art CNN compression methods. The evaluated model architectures are four widely-used networks, ResNet-56 \cite{he2016deep}, WRN-40-1 \cite{zagoruyko2016wide}, DenseNet-40-12 \cite{huang2017densely} and MobileNet V2 \cite{sandler2018mobilenetv2}. The results are summarized in Table \ref{tab5}.

\textbf{ResNet-56.}
We compare the proposed OD method with state-of-the-art structured pruning methods and compact convolutions. ResNet-56-P2/P4 \cite{singh2019hetconv} is a compact network with the heterogeneous kernel-based convolution. As shown in Table \ref{tab5},
compared to all structured pruning methods \cite{he2017channel,li2017pruning,yu2018nisp,he2018amc,he2018soft,ding2019centripetal,he2019filter,lin2019towards}, ResNet-56-P2 achieves the lower error of 6.40\%, but still has high computation cost. Through increasing the part $P$, which controls the number of different types of kernels in a convolutional filter, the FLOPs can be reduced significantly. This is just like the number of group in group convolution.
Instead, C-SGD achieves a comparable result with ResNet-56-P4, and the best trade-off between error and FLOPs, compared to other structured pruning methods. 
When group convolution is equipped into OD, the proposed OD-Group-16 achieves the best performance, compared to all the state-of-the-art methods.
For example, OD-Group-16 achieves a factor of $5.3\times$ compression rate and $5.43\times$ speedup rate, and accuracy improvement of 0.08\%, compared to the baseline.

\textbf{WRN-40-1.}
WRN-40-1 contains 40 layers with widening factor $k=1$, which achieves 6.46\% error with the number of 0.56M parameters and 83.02M FLOPs.
In \cite{crowley2018moonshine}, WRN-40-2 is used to be a teacher model, which is significantly larger than WRN-40-1.
To reduce the number of parameters in WRN-40-2, Crowley \emph{et al.} replace the standard convolution with the depthwise convolution, termed WRN-40-2-Depthwise, and employ attention transfer to improve the performance. Our method is typically different by using online distillation without a pre-trained teacher model. Moreover, compared to WRN-40-2-Depthwise, the proposed OD-Group-16 requires significantly fewer parameters (0.1M \emph{vs.} 0.29M), while increases the error with only 0.21\% (\emph{i.e.}, 6.71\% \emph{vs.} 6.5\%).
 
\textbf{DenseNet-40-12.}
we further evaluate the performance of the proposed OD with cheap convolutions. As shown in Table \ref{tab5}, Liu \emph{et al} \cite{liu2017learning} proposed iterative network slimming method, which achieves the lowest error of 5.19\%, but only prunes the number of 37.74\% parameters (\emph{i.e.}, 0.66M parameters used). Compared to Liu \emph{et al} \cite{liu2017learning}, GAL \cite{lin2019towards} and C-SGD \cite{ding2019centripetal}, shift convolution is merged into OD, which substantially reduces the parameters. Therefore, it achieves the highest compression rate and speedup rate by 4.24$\times$ and 3.54$\times$, respectively. 
For depthwise convolution, OD-depthwise achieves a comparable result with OD-shift. For example, OD-Depthwise achieves a lower error of 5.79\% (\emph{vs.} 6.41\% in OD-Shift), by a lower compression rate of 3.21$\times$ (\emph{vs.} 4.24$\times$ in OD-Shift).
Therefore, we can select the suitable models according to the requirement of practical applications.   

\textbf{MobileNet V2.}
The proposed OD can also be used to improve the performance of compact networks. We take the MobileNet V2 \cite{sandler2018mobilenetv2} for example. The error of MobileNet V2 model is obviously reduced by training with OD. This indicates that mutual learning in OD can improve the performance of both student models and teacher model.


\subsubsection{ImageNet}
We also evaluate the proposed CNN compression scheme on ImageNet-1K using ResNet-18 and ResNet-50. Each Residual block in ResNet-18 contains two $3\times 3$ standard convolutions and shortcut connections, while the bottleneck structure is used in ResNet-50, which contains shortcut connections and three standard convolutional layers with $1\times 1, 3\times 3$ and $1\times 1$ convolutions. Each convolution is followed by both batch normalization and ReLU. For simplicity and friendly implementation, we replace the $3\times 3$ convolution with cheap convolutions and keep the $1\times1$ convolution\footnote{It does not mean that the standard $1\times1$ convolution cannot be compressed by the cheap convolutions. We keep it for better implementation.}. We summarize the results for ResNet-18 and ResNet-50 compression in Table \ref{tab6}.

\begin{table*}[!t]
\footnotesize
\begin{center}
\caption{Results of compressing ResNet-18 and ResNet-50 on ImageNet.}
\label{tab6}
\begin{tabular}{cccccc}
\hline
\multirow{2}*{Model} & \multirow{2}*{Method} & \multicolumn{1}{c}{Top-1}  & \multicolumn{1}{c}{Top-5} & \multirow{2}*{\#Params (M)} & \multirow{2}*{FLOPs (B)} \\
&  & error \%&  error \%&  & \\\Xhline{0.1em}
\hline
\multirow{7}{*}{ResNet-18} & Baseline & 29.72 & 10.37 & 11.69 & 1.81 \\ 
& SFP \cite{he2018soft} & 32.90 & 12.22 & - & 1.06 \\
& PFGM \cite{he2019filter} & \textbf{31.59} & 11.52 & - & 1.06 \\
& OD-Shift & 39.22 & 16.53 & \textbf{1.92} & \textbf{0.32} \\
& OD-Depthwise & 35.73 & 12.83 & 1.96 & 0.34 \\
& OD-Group-2 & 31.65 & \textbf{11.33} & 7.03 & 1.12\\
& OD-Group-4 & 34.06 & 12.78 & 4.48 & 0.72 \\
\hline \hline
\multirow{15}{*}{ResNet-50} & Baseline & 23.85 & 7.13 & 25.56 & 4.09 \\ 
& ThiNet-50 \cite{luo2017ThiNet} & 28.99 & 9.98 & 12.38 & 1.71 \\
& ThiNet-30 \cite{luo2017ThiNet} & 31.58 & 11.70 & \textbf{8.66} & \textbf{1.10} \\
& He \emph{et al.} \cite{he2017channel} & 27.70 & 9.20 & - & 2.73 \\
& GDP-0.6 \cite{lin2018accelerating} & 28.81 & 9.29 & - & 1.88 \\
& GDP-0.5 \cite{lin2018accelerating} & 30.42 & 9.86 & - & 1.57 \\
& SFP \cite{he2018soft} & 25.39 & \textbf{7.13} & - & 2.42\\
& SSS-32 \cite{huang2018data} & 25.82 & 8.09 & 18.6 & 2.82 \\
& SSS-26 \cite{huang2018data} & 28.18 & 9.21 & 15.6 & 2.33 \\
& GAL-0.5 \cite{lin2019towards} & 28.05 & 9.06 & 21.2 & 2.33 \\
& GAL-1 \cite{lin2019towards} & 30.12& 10.25 & 14.67 & 1.58 \\
& ResNet-50-P4 \cite{singh2019hetconv} & \textbf{23.84} & - & - & 2.85 \\
& OD-Shift & 28.81 & 9.21 & 14.23 & 2.00 \\
& OD-Depthwise & 26.37 & 7.34 & 14.27 & 2.26 \\
& OD-Group-16 & 26.54 & 7.53 & 14.95 & 2.42 \\
\hline
\end{tabular}
\end{center}
\end{table*}

\textbf{ResNet-18.}
Compared to SFP \cite{he2018soft} and PFGM \cite{he2019filter}, the proposed OD-Shift achieves the highest speedup by a factor of 5.66$\times$ (\emph{i.e.}, 0.32B FLOPs \emph{vs.} 1.06B FLOPs in both SFP and PFGM).
However, OD-Shift leads to a significant accuracy drop. To improve the accuracy, we can replace the standard convolution with group or depthwise convolution. For example, OD-Depthwise achieves the 3.49\% improvement in accuracy, but only with an increase of 0.02B in FLOPs, compared to shift convolution. 
In addition, OD-Group-2 achieves a comparable result with PFGM, \emph{i.e.}, 11.33\% Top-5 error with 1.12B FLOPs in OD-Group-2 \emph{vs.} 11.53\% Top-5 error with 1.06B FLOPs in PFGM.

\textbf{ResNet-50.} 
As shown in Table \ref{tab6}, ResNet-50-P4 \cite{singh2019hetconv} achieves the lowest Top-1 error of 23.84\% by using the heterogenous convolution. However, it requires the highest FLOPs of 2.85B to classify an image with a size of $224\times 224$. 
For ThiNet-30 \cite{luo2017ThiNet}, the number of parameters and FLOPs can be substantially reduced by layer-wise compression and error reconstruction. However, it leads to a highest increase of 7.73\% error, which is due to the removal of some important filters.
Several methods consider the better trade-off between error and parameters/FLOPs, such as GDP \cite{lin2018accelerating}, SSS \cite{huang2018data} and GAL \cite{lin2019towards}. 
Compared to SSS-26 and GAL-0.5, the proposed OD-Depthwise achieves the best performance by a factor of 1.79$\times$ compression (\emph{vs.} 1.64$\times$ and 1.21$\times$ compression in SSS-26 and GAL-0.5, respectively), with a lowest Top-1/5 error of 26.37\%/7.34\% (\emph{vs}. Top-1/5 error of 28.18\%/9.21\% and 28.05\%/9.06\% in SSS-26 and GAL-0.5, respectively). 
Note that different cheap convolutions have less effect on their parameters, but can affect the classification error. 

\section{Conclusion}
\label{sec5}
In this paper, we propose a novel CNN compression scheme by replacing the standard convolution with the cheap one without redesigning the network architecture, and using knowledge distillation to further improve the performance of compressed networks. In particular, we propose online distillation method to conduct mutual learning between multiple students and a strong online teacher without stage-wise optimization. We conduct comprehensive experiments to evaluate the effectiveness of the proposed method on a variety of CNN architectures over different datasets. 

\section{Acknowledgements}

\section*{References}

\bibliography{nc}

\end{document}